\documentclass{article}

\usepackage[final]{nips_2018}

\usepackage[utf8]{inputenc} 
\usepackage[T1]{fontenc}    
\usepackage{hyperref}       
\usepackage{url}            
\usepackage{booktabs}       
\usepackage{amsfonts}       
\usepackage{nicefrac}       
\usepackage{microtype}      
\usepackage{natbib}
\usepackage{xcolor}
\usepackage{graphicx}
\usepackage{subcaption}     
\setcitestyle{round}
\usepackage{wrapfig}

\usepackage{amsmath}
\def\argmax#1{\mathrm{arg\ max}_{#1}\,}
\def\max#1{\mathrm{max}_{#1}\,}

\newtheorem{hypothesis}{Hypothesis}

\title{Deep Reinforcement Learning and the Deadly Triad}

\author{
  Hado van Hasselt \\DeepMind \And
  Yotam Doron \\DeepMind \And
  Florian Strub \\University of Lille \\ DeepMind \AND
  Matteo Hessel \\DeepMind \And
  Nicolas Sonnerat \\DeepMind  \And
  Joseph Modayil \\DeepMind 
}

\begin{document}

\maketitle

\begin{abstract}
We know from reinforcement learning theory that temporal difference learning can fail in certain cases.  \citet{SuttonBarto:2018} identify a \emph{deadly triad} of function approximation, bootstrapping, and off-policy learning.  When these three properties are combined, learning can diverge with the value estimates becoming unbounded.
However, several algorithms successfully combine these three properties, which indicates that there is at least a partial gap in our understanding.
In this work, we investigate the impact of the deadly triad in practice, in the context of a family of popular deep reinforcement learning models--- deep Q-networks trained with experience replay---analysing how the components of this system play a role in the emergence of the deadly triad, and in the agent's performance.
\end{abstract}

In this paper, we focus on temporal difference (TD) methods \citep{Sutton:1988,SuttonBarto:1998,SuttonBarto:2018} to learn value functions. The value of a state, or of an action, is the expected discounted return that would result from following a particular behaviour policy from that point onward.  An {\em off-policy} learning algorithm can estimate values for a policy that differs from the actual behaviour, which allows an agent to learn about policies other than the one it follows and to learn about many different policies in parallel \citep{Sutton:2011}.  It also allows an efficient form of policy improvement \citep{Howard:1960} by directly estimating the values of a policy that is greedy with respect to current value estimates, as used in value iteration \citep{Bellman:57,Howard:1960,Puterman:1994} and Q-learning \citep{Watkins:1989,WatkinsDayan:92}.

In the simplest form of TD learning, the immediate reward is added to the discounted value of the subsequent state, and this is then used as a target to update the previous state's value. This means the value estimate at one state is used to update the value estimate of a previous state---this is called \emph{bootstrapping} \citep{SuttonBarto:2018}.  Bootstrapping is also commonly used within policy-gradient and actor-critic methods to learn value functions  \citep{Witten:1977,BartoSuttonAnderson:83,Peters:2008NAC,bhatnagar2009natural,vanHasselt:2012,Degris:2012,Mnih:2016}.

For many large problems, learning the value of each state separately is not possible. In this common situation, value function approximation must generalise across states.  When combining TD learning with function approximation, updating the value at one state creates a risk of inappropriately changing the values of other states, including the state being bootstrapped upon.  This is not a concern when the agent updates the values used for bootstrapping as often as they are used~\citep{Sutton:2016}. However, if the agent is learning off-policy, it might not update these bootstrap values sufficiently often. This can create harmful learning dynamics that can lead to divergence of the function parameters \citep{Sutton:1995,Baird:1995,Tsitsiklis:1997}.  The combination of function approximation, off-policy learning, and bootstrapping has been called ``the deadly triad'' due to this possibility of divergence~\citep{SuttonBarto:2018}. 

Partial solutions have been proposed \citep{Baird:1995,Precup:2001,Sutton:2008,Maei:2009,Maei:2011,Sutton:2016}, but they have mostly not been extended to non-linear function approximation, and have not been widely adopted in practice. As mentioned by Sutton and Barto (2018, Chapter 11.10),
 ``\emph{The potential for off-policy learning remains tantalizing, the best way to achieve it still a mystery}''.

On the other hand, perhaps surprisingly, many algorithms successfully combine all components in the deadly triad.  The deep Q-network (DQN) agent proposed by \citet{Mnih:2013,Mnih:2015} uses deep neural networks to approximate action values, which are updated by Q-learning, an off-policy algorithm.  Moreover, DQN uses experience replay \citep{Lin:1992} to sample transitions, thus the updates are computed from transitions sampled according to a mixture of past policies rather than the current policy. This causes the updates to be even more off-policy. Finally, since DQN uses one-step Q-learning as its learning algorithm, it relies on bootstrapping. Despite combining all these components of the deadly triad, DQN successfully learnt to play many Atari 2600 games \citep{Mnih:2015}.

In this paper, we conduct an empirical examination into when the triad becomes deadly.  In particular, we investigate the space of algorithms around DQN to see which of these variants are stable and which are not, and which factors lead to learning that is reliable and successful.

\section{The deadly triad in deep reinforcement learning}

If the triad is deadly, one might ask why does DQN work? To examine this question more carefully, we first note that each of the components of the triad is non-binary, and can be modulated as follows:

\paragraph{Bootstrapping}
We can modulate the influence of bootstrapping using multi-step returns \citep{Sutton:1988,SuttonBarto:2018}, as increasing the number of steps before bootstrapping reduces its impact. Multi-step returns have already been shown to be beneficial for performance in certain variations of DQN \citep{Hessel:2018, Horgan:2018}. 

\paragraph{Function approximation}
We can modify the generalisation and aliasing of the function approximation by changing the capacity of the function space. We manipulated this component by changing the size of the neural networks.

\paragraph{Off-policy}
Finally, we can change how off-policy the updates are by changing the state distribution that is sampled by experience replay. In particular, we can do so by prioritising certain transitions over others \citep{Schaul:2016}. Heavier prioritisation can lead to more off-policy updates.

By systematically varying these components, we can investigate when algorithms become unstable in practice, and gain insights into the relationships between the components of the triad and the learning dynamics. In particular, we can observe under which conditions instabilities appear during training.

As a first step, we examine the deadly triad in a well-known toy example, where we can 
analyse the learning dynamics of our algorithms, to build intuition about the problem.  We then thoroughly investigate the properties of the deadly triad in Atari 2600 games from the Arcade Learning Environment \citep{Bellemare:2013}, to ensure that our conclusions do not only apply to simple examples that are perhaps contrived, but also to much larger domains.

\section{Building intuition}

\def\w{\mathbf{w}}
The goal of TD learning is to update the weights $\w$ of a value function $v_{\w}$ to make it close to the true value $v_{\pi}$ of policy $\pi$ that defines the (possibly stochastic) mapping from states to actions $A_t \sim \pi(S_t)$:
\[
v_{\w}(s)
~~\approx~~ v_{\pi}(s)
~~=~~ \mathbb{E}\left[ R_{t+1} + \gamma R_{t+2} + \ldots \mid S_t = s, A_{t+i} \sim \pi(S_{t+i})\,,\forall i > 0 \right] \,.
\]
TD learning \citep{Sutton:1988} achieves this with an update
\begin{equation}
\label{td}    
\Delta\w ~~\propto~~
(R_{t+1} + \gamma v_{\w}(S_{t+1}) - v_{\w}(S_t)) \nabla_{\w} v_{\w}(S_t) \,,
\end{equation}
where $v_{\w}(S_{t+1})$ is used to estimate the unknown remainder of the return after state $S_{t+1}$.

\begin{figure}
\centering
\begin{subfigure}[t]{.32\textwidth}
\raisebox{0.1\height}{\includegraphics[width=.95\textwidth]{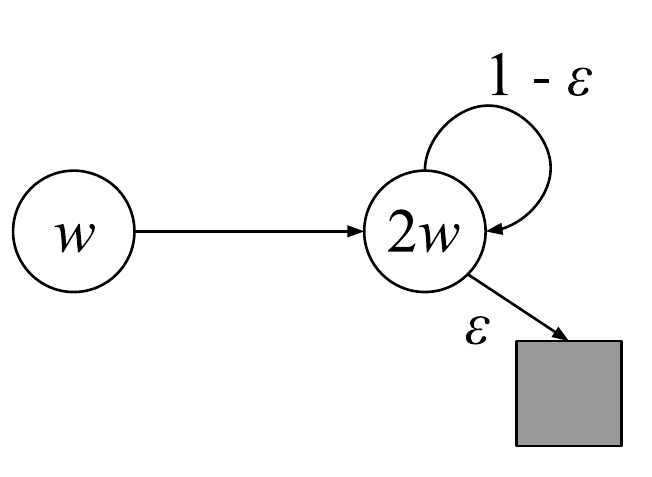}}
\caption{\small \label{fig:triadexample} The example by \citet{Tsitsiklis:1997}.}
\end{subfigure}
~
\begin{subfigure}[t]{.32\textwidth}
\includegraphics[width=.96\textwidth]{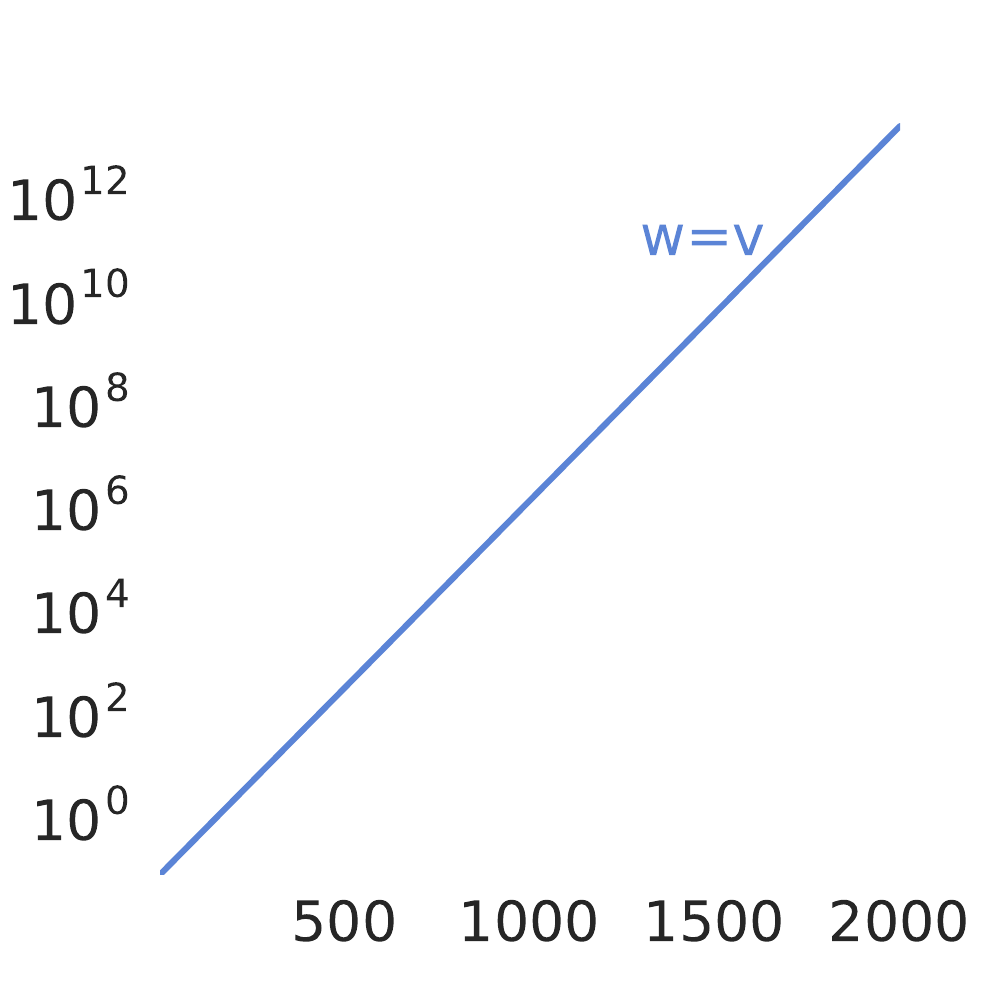}
\caption{\small \label{fig:divergence} $v(s) = w \phi(s)$ diverges.}
\end{subfigure}
~
\begin{subfigure}[t]{.32\textwidth}
\includegraphics[width=.96\textwidth]{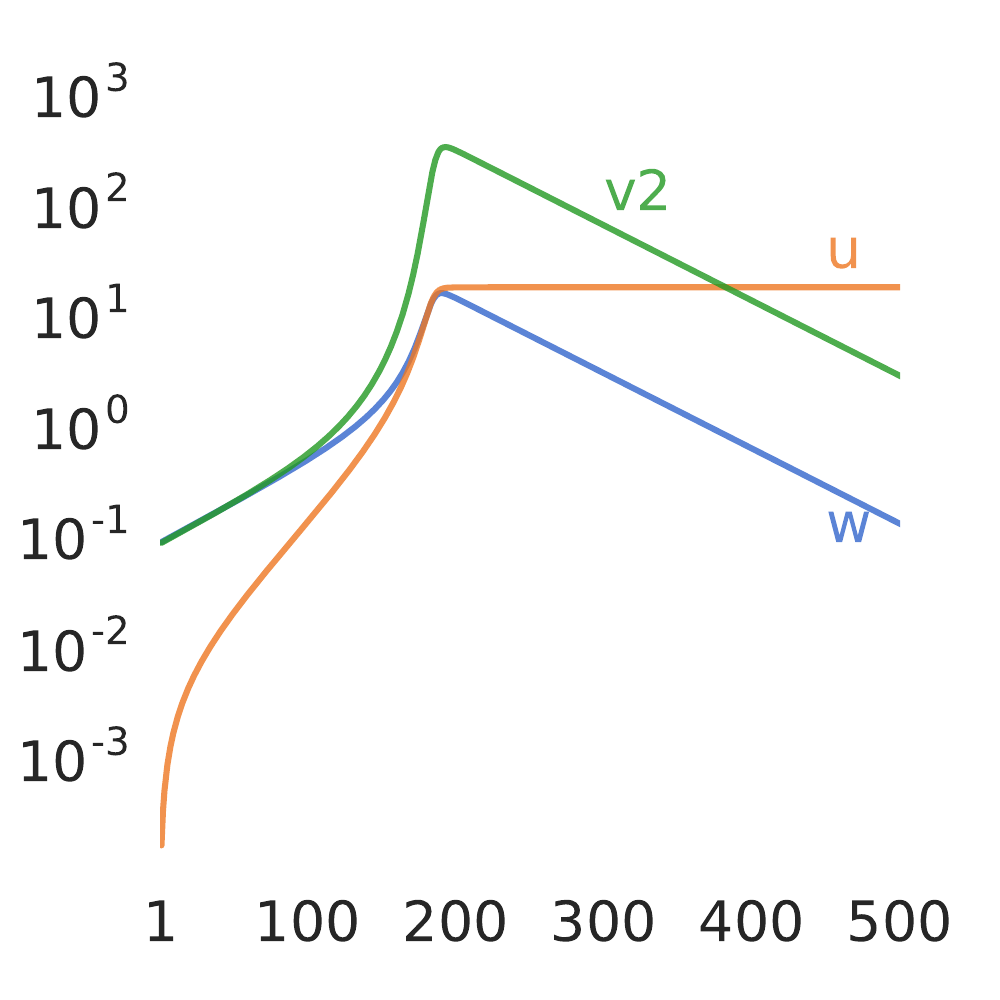}
\caption{\small \label{fig:convergence} $v(s)\!=\!w(\phi(s)\!+\!u)$ converges.}
\end{subfigure}
\caption{\label{fig:w2w} When sampling each state equally often, the example \ref{fig:triadexample} diverges with $v(s) = w \phi(s)$ (\ref{fig:divergence}).  However, it converges (\ref{fig:convergence}) when we introduce one additional parameter, $v(s) = w(\phi(s)+ u)$.  Both systems are initialized to the same point (with $u=0$) and optimized identically.  The number of learning updates are shown on the x-axis, and the values are shown on the y-axis in log-scale.}
\end{figure}

Several examples of divergence have been proposed in literature \citep{Baird:1995, Tsitsiklis:1997, Sutton:2016}. Next, we illustrate the nature of the problem using the example by \citet{Tsitsiklis:1997}, shown in Figure \ref{fig:triadexample}.
In this example, each state is described by a single scalar feature $\phi$, such that $\phi(s_1) = 1$ and $\phi(s_2) = 2$. The estimated value in each state is $v(s) = w \times \phi$, where $w$ is the weight we will update.  In particular, we then have $v(s_1) = w$ and $v(s_2) = 2w$. These value estimates are shown in the circles in the figure.  All rewards are $0$, and therefore the value predictions would be perfect with $w_* = 0$.

If we update the value in each state according to the dynamics (i.e., \emph{on-policy}), then the value of state $s_2$ is updated, in expectation, multiple times for each update to the value of state $s_1$. Then, the weight $w$ will converge to its optimal setting, $w_* = 0$.

Now consider updating only the value of $s_1$, but not the value of $s_2$. This is an off-policy update, since the frequency of updates for the value of $s_2$ does not match the actual frequency of visiting $s_2$ under the dynamics of the problem. The TD update to $w$, according to \eqref{td} is then
$
\Delta w \propto
\gamma 2 w - w =
(2\gamma -1) w
$. 
For a discount $\gamma > 1/2$, we have $2 \gamma  > 1$ and so any weight $w \neq 0$ will be updated away from zero, regardless of its current value: $w$ diverges.  Even if we update $s_2$ as often as $s_1$, this would be less often than under on-policy updating, and we will still see divergence for large enough $\gamma$ (Figure \ref{fig:divergence}).

Intuitively, the problem highlighted by this example is that when we update certain state values, we might update other values as well, due to the generalisation introduced by function approximation. If, in addition, we are learning off-policy, we might not correct for these inadvertent updates sufficiently, and parameters and value estimates might therefore spiral out of control.

We note, however, that the example can converge with a small modification to the function class.  We define the state values to be $v(s_i) = w( \phi(s_i) + u)$, where $u$ is also learnable.  If we fix $u=0$, we recover the canonical example.  But, if we allow both $w$ and $u$ to be updated by TD, then values first seemingly diverge, then recover and converge to the optimum $v(s_1) = v(s_2) = 0$ (see Figure \ref{fig:convergence}).

Our understanding of divergence under function approximation is limited.  Linear function approximation for reinforcement learning is extensively studied, but the examples of divergence are typically contrived. 
Furthermore, when using deep non-linear function spaces with reinforcement learning, there is little guidance on whether divergence is common, and if the deadly triad is to blame.

\section{Hypotheses}\label{HypSect}
In the remainder of the paper, we investigate several concrete hypotheses about how different algorithmic components contribute to the divergence of the learning process.  Our starting point is a fairly standard variant of the DQN algorithm, to allow our conclusions to be informative about real current practice rather than being specific only to carefully crafted edge cases.
To test our hypotheses, we investigated many variants of DQN, trained with different hyper-parameters, and systematically tracked where divergence occurred.

\begin{hypothesis}[Deep divergence]\label{hyp:vanilla}
Unbounded divergence is uncommon when combining Q-learning and conventional deep reinforcement learning function spaces.
\end{hypothesis}
Motivated by the number of successful applications of deep reinforcement learning, this hypothesis is a deep-network analogue to empirical results with linear function approximation.  Although divergence can theoretically occur from the off-policy nature of Q-learning with an $\epsilon$-greedy policy, it is rare in practice~\citep[Section 11.2]{SuttonBarto:2018}. We tested whether the same is true in our set of deep reinforcement learning experiments.

The following two hypothesis refer to variants of the Q-learning update rule.  These do not correspond directly to the components of the triad, which are covered later, but they may still interact with the learning dynamics to make divergences more or less probable. All updates have the form
\[
\Delta\theta \propto (G^{(n)}_t - q(S_t, A_t))\nabla_{\theta} q(S_t, A_t) \,,
\]
where $t$ is a time step from the experience replay and $G^{(n)}_t = R_{t+1} + \ldots + \gamma^{n-1} R_{t+n} + \gamma^n v(S_{t+n})$ is a $n$-step return. For instance, for standard one-step Q-learning, $v(s) = \max{a} q(s, a)$ and $n=1$.

\begin{hypothesis}[Target networks]\label{hyp:network}
There is less divergence when bootstrapping on separate networks.
\end{hypothesis}
\citet{Mnih:2015} showed that it can be useful to bootstrap on $v(s) = \max{a} q'(s, a)$, where $q'$ is a sporadically updated separate copy of the online network $q$. We call this \emph{target Q-learning}.  In the context of the deadly triad, it makes sense that this could be beneficial, as the bootstrap target can not be inadvertently updated immediately if a separate network is used.  However, target networks do not suffice as a solution to the deadly triad.   When such target networks are applied to the standard Tsitsiklis and Van Roy example (Figure~\ref{fig:triadexample}) with linear function approximation, the weights still diverge, though the divergence is slowed down by the copying period.

\begin{hypothesis}[Overestimation]\label{hyp:bias}
There is less divergence when correcting for overestimation bias.
\end{hypothesis}
Standard Q-learning and target Q-learning are known to suffer from an overestimation bias \citep{vanHasselt:2010, vanHasselt:2016}. To prevent this, we can decouple the action selection from the action evaluation in the bootstrap target, by using $v(s) = q'(s, \argmax{a} q(s, a))$.  This is known as double Q-learning \citep{vanHasselt:2010}.

Double Q-learning as defined above, and as combined with DQN by \citet{vanHasselt:2016}, uses the slow-moving copy $q'$ to evaluate the selected action. This merges the benefits of reducing overestimation (by decoupling the selection and evaluation of the action), with those of using a separate stationary target network.  To disentangle these effects, we can define a new double Q-learning variant, which bootstraps on $v(s) = q(s, \argmax{a} q'(s, a))$ when updating $q$.  We call this \emph{inverse double Q-learning}. It uses the same network, $q$, as is being updated to obtain value estimates, but a separate target network to determine the action. Therefore, it has the benefits of reducing overestimation, but not those of using a separate target network for bootstrapping.

In summary, we obtain four different bootstrap targets:
\begin{align*}
v(s) & = q(s, \argmax{a} q(s, a)) \phantom{''}= \max{a} q(s, a) \tag{Q-learning} \\
v(s) & = q'(s, \argmax{a} q'(s, a)) = \max{a} q'(s, a) \tag{target Q-learning} \\
v(s) & = q(s, \argmax{a} q'(s, a)) \tag{inverse double Q-learning}\\
v(s) & = q'(s, \argmax{a} q(s, a)) \tag{double Q-learning}\,.
\end{align*}
If hypothesis \ref{hyp:network} is true, we would expect (target Q $\prec$ Q) and (inverse double Q $\prec$ double Q), where we define `$\prec$' loosely to mean `diverges less often than'.  Hypothesis \ref{hyp:bias} would imply (inverse double Q $\prec$ Q) and (double Q $\prec$ target Q). Together, these would then define a partial ordering in which Q-learning diverges most, and double Q-learning diverges least, with the other variants in between.

\begin{hypothesis}[Multi-step]\label{hyp:steps}
Longer multi-step returns will diverge less easily.
\end{hypothesis}
We can use multi-step returns \citep{SuttonBarto:2018} to reduce the amount of bootstrapping. When we bootstrap immediately, after a single step, the contraction in the (linear or tabular) learning update is proportional to the discount $\gamma \in [0, 1]$.  When we only bootstrap after two steps the update may be noisier, but the expected contraction is $\gamma^2$. Intuitively, we might diverge less easily when using multi-step updates (using larger $n$ in the definition of $G^{(n)}_t$), as we bootstrap less.
This hypothesis is supported with linear function approximation, but, since a TD-operator applied to non-linear deep function spaces is not a formal contraction on the parameter space, it is worth checking if this intuition still holds true when using deep networks to estimate values.

\begin{hypothesis}[Capacity]\label{hyp:capacity}
Larger, more flexible networks will diverge less easily.
\end{hypothesis}
One part of the problem is the inappropriate generalization across states. 
If all values are stored independently in a function approximator, then divergence would not happen---this is why the tabular version of off-policy TD does not diverge.  We hypothesize that more flexible function approximation might behave more like the tabular case, and might diverge less easily.

\begin{hypothesis}[prioritisation]\label{hyp:pri}
Stronger prioritisation of updates will diverge more easily.
\end{hypothesis}
Most counterexamples modify the state distribution to induce divergence, for instance by updating all states equally often while the on-policy distribution would not \citep{Baird:1995, Tsitsiklis:1997}.  To modify this distribution, we can use prioritised replay \citep{Schaul:2016}.
Specifically, the probability $p_k$ of selecting state-action pair $(S_k, A_k)$ is a function of the TD-error
\[
p_k \propto \lvert G^{(n)}_k - q(S_k, A_k) \rvert^\alpha \,,
\]
where $q(S_k, A_k)$ and the multi-step return $G^{(n)}_k$ are the respective values when the value of state-action pair $(S_k, A_k)$ was first put into the replay, or when this sample was last used in an update.
For $\alpha=0$, we obtain uniform random replay.  For $\alpha>0$, the sampling distribution can be corrected toward uniform by multiplying the resulting update with a importance-sampling correction $1/(N p_k)^\beta$, where $N$ is the size of the replay.  We can modulate how off-policy the updates are with $\alpha$ and $\beta$.

\section{Evaluation}

To examine these hypotheses empirically, we ran a variety of experiments using the Atari Learning Environment~\citep{Bellemare:2013} using variants of DQN~\citep{Mnih:2015}.  We used the same preprocessing as \citep{Mnih:2015}---details are given in the appendix.

The following parameters were systematically varied in our main experiments.
The algorithmic parameters include choosing one of the four \textbf{bootstrap targets} described in the previous section, and the \textbf{number of steps before bootstrapping} with $n=1$, $n=3$, or $n=10$.
We tested four levels of \textbf{prioritisation} corresponding to $\alpha\in\{0, \frac{1}{2}, 1, 2\}$, both with ($\beta=0.4$, as proposed by \citet{Schaul:2016}) and without ($\beta=0$) importance sampling corrections.  Finally, we tested with four different \textbf{network sizes}, referred to as \textit{small}, \textit{medium}, \textit{large} and \textit{extra-large} (details are given in the appendix).

These parameter configurations gives a total of 336 parameter settings per game, for 57 different games, which jointly cover a fairly large space of different configurations and problems. Each configuration was run for roughly 20M frames on a single CPU core (which requires a few days per configuration).  The duration of each run is not enough for state-of-the-art performance but it is sufficient to investigate the learning dynamics, and it allows us to examine a broad set of configurations. For reliable results, we ran 3 distinct replications of each experiment.

We tracked multiple statistics during these experiments, over roughly 50K--100K frame intervals.  The most important statistic we tracked was the maximal absolute action value estimate, denoted ``maximal |Q|'' in the figures. We use this statistic to measure stability of the value estimates. Because the rewards are clipped to $[-1, 1]$, and because the discount factor $\gamma$ is $0.99$, the maximum absolute true value in each game is bounded by $1 + \gamma + \gamma^2 + \ldots = \frac{1}{1 - \gamma} = 100$ (and realistically attainable values are typically much smaller).  Therefore, values for which $|q| > 100$ are unrealistic. We call this phenomenon \emph{soft divergence}.

\subsection{Unbounded divergence in deep RL (Hypothesis \ref{hyp:vanilla})}

\begin{figure}
\centering
\includegraphics[width=.44\linewidth]{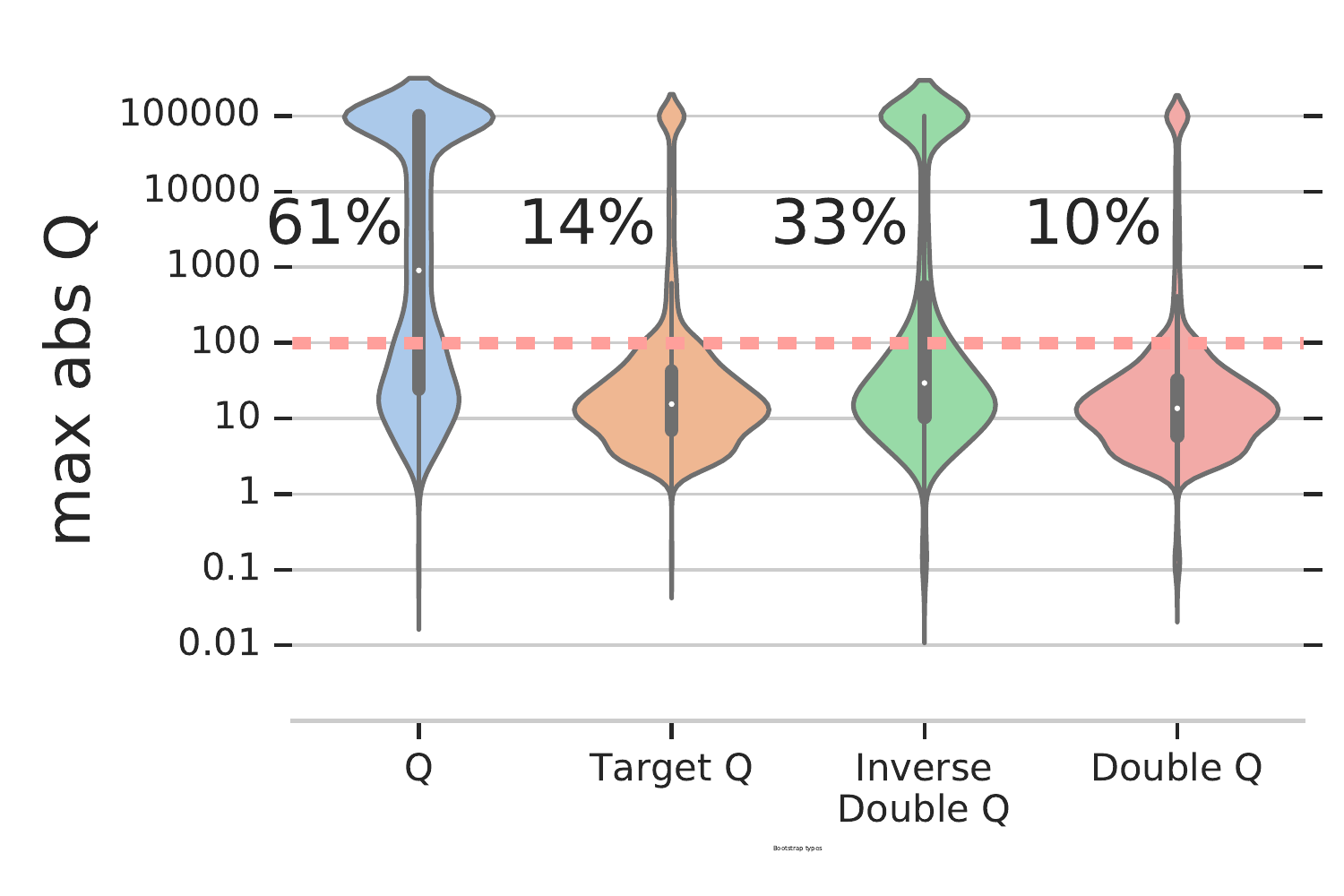}
\includegraphics[width=.55\textwidth]{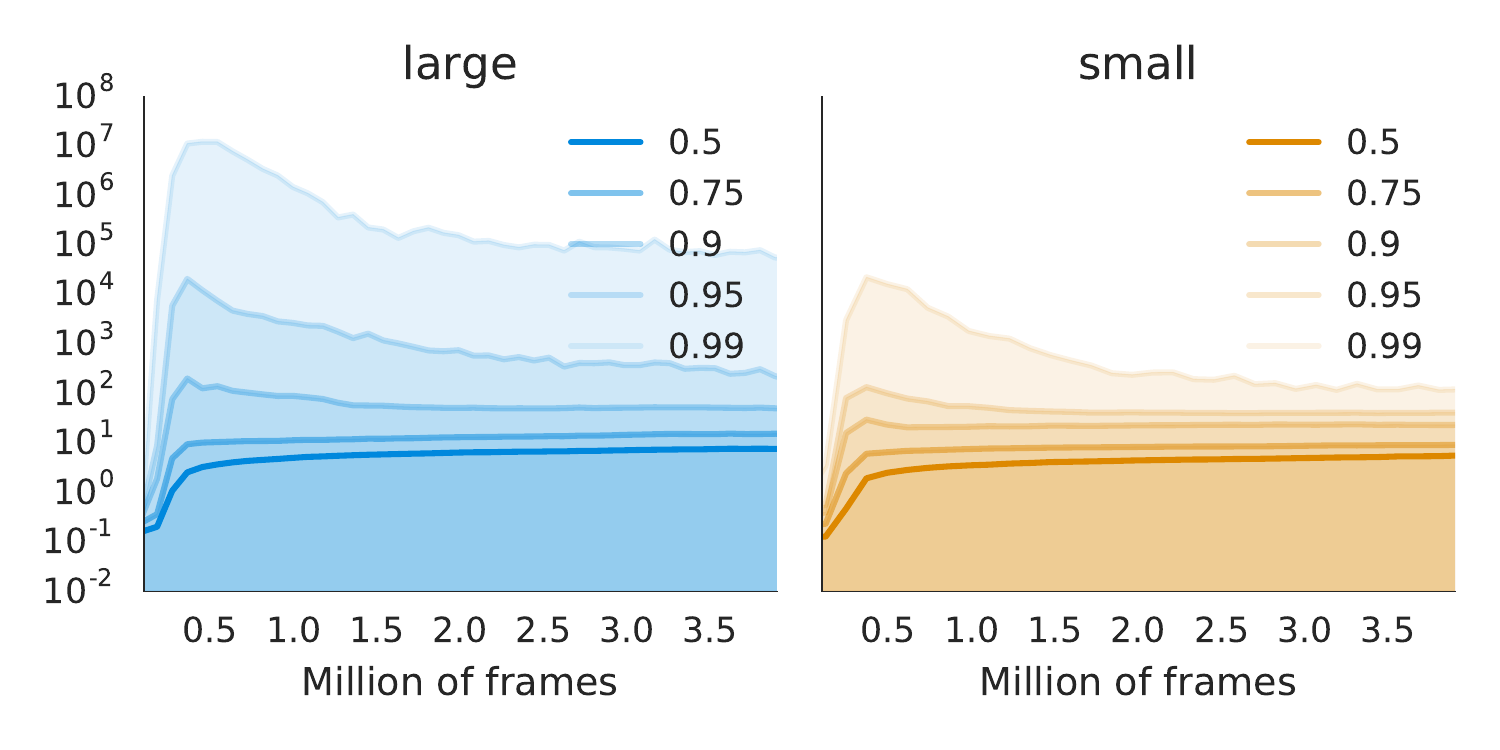}
\caption{{\bf Maximal absolute action-values distribution.} 
On the left, we report the number of occurrences of soft-divergence (values beyond the maximum realisable range of $[-100, 100]$), split by bootstrap type.
On the right, we plot the percentiles of the maximal absolute Q-value, on a logarithmic scale, for two different network size. The values on many runs quickly rises to unrealistic magnitudes, but then shrinks down to a plausible magnitude.
}\label{fig:vanilla}
\end{figure}

We first examined whether the values sometimes diverge, as in the canonical examples of the deadly triad.
The results from these trials (Figure~\ref{fig:vanilla}) show that soft divergences do occur (values exceed $100$), and sometimes grow to wildly unrealistic value estimates.  But, surprisingly, they never became unbounded (more precisely, they never produced floating point NaNs).
The lack of unbounded divergence in any of the runs suggests that although the deadly triad can cause unbounded divergence in deep RL, it is not common when applying deep Q-learning and its variants to current deep function approximators. This provides support for Hypothesis~\ref{hyp:vanilla}.

\subsection{Examining the bootstrap targets (Hypothesis \ref{hyp:network}, \ref{hyp:bias})}

As unbounded divergence can be rare, we now focus on occurrences of soft-divergence. For each of $336$ parameter settings and 3 replications, we tracked maximal absolute Q-value across time steps.

First, we examined the sensitivity of different statistical estimation techniques to the emergence of soft-divergence (Hypothesis \ref{hyp:network}, \ref{hyp:bias}). Specifically, we examined the stability of the four update rules from Section \ref{HypSect}: Q-learning, target Q-learning, inverse double Q-learning and double Q-learning. 
Figure~\ref{fig:vanilla}, on the left hand side, shows that Q-learning exhibits by far the largest fraction of instabilities (61\%); target Q-learning and double Q-learning, which both reduce the issue of inappropriate generalization via the use of target networks, are the most stable. Inverse double Q-learning, which addresses the over-estimation bias of Q-learning but does not benefit from the stability of using a target network, exhibits a moderate rate (33\%) of soft-diverging runs. These results provide strong support for both Hypothesis~\ref{hyp:network} and \ref{hyp:bias}: divergence occurs more easily with overestimation biases, and when bootstrapping on the same network.

To further understand the nature of soft-divergence, we tracked the maximum absolute values over time. We observed that while value estimates frequently grew quickly to over a million, they typically reduced down again below the threshold of 100. On the right of Figure~\ref{fig:vanilla}, we plot the distribution of maximum absolute action values across runs, as a function of time, for two different network sizes. This reproduces at scale the pattern seen for the modified \cite{Tsitsiklis:1997} example, when more learnable parameters are added to the canonical formulation: values initially grow very large, but then return to more accurate value estimates.

\begin{figure*}[t]
    \centering
    \includegraphics[width=0.975\textwidth]{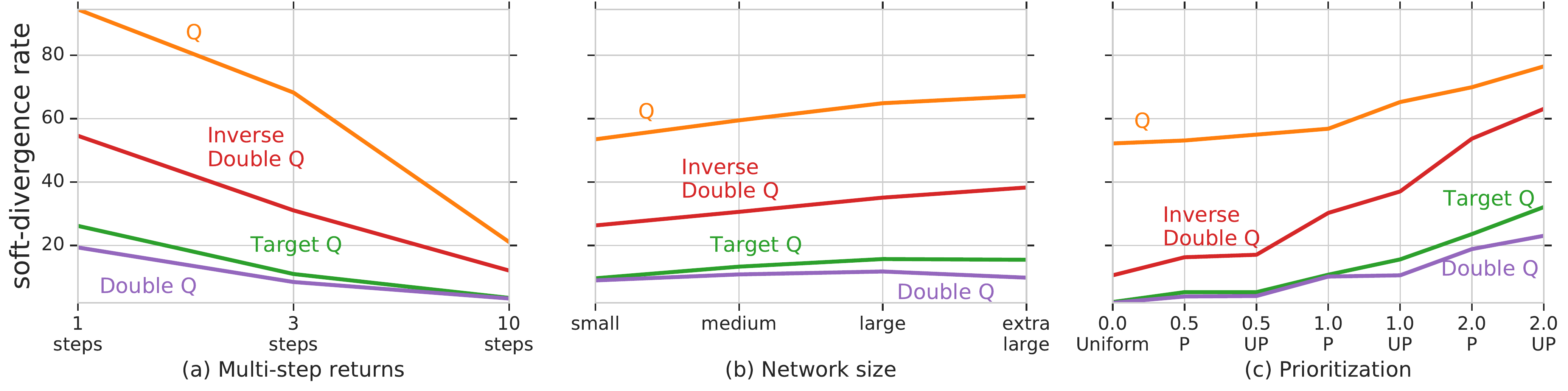}
    \caption{\textbf{Soft-divergence} We plot, for each bootstrap type, the fraction of runs that exhibit soft-divergence (values beyond the maximum realisable range of $[-100, 100]$), respectively as a function of bootstrap length, network size, and type of prioritisation. Percentages are across all other parameter configurations, and 3 replications. (a) Increasing the bootstrap length decreases instability (supports Hypothesis \ref{hyp:steps}). (b) Increasing the network capacity increases instability (against Hypothesis~\ref{hyp:capacity}). (c) Increasing the prioritisation increases instability (supports Hypothesis~\ref{hyp:pri}).}\label{fig:violin_triad}
\end{figure*}

\subsection{Examining the deadly triad (Hypotheses \ref{hyp:steps}, \ref{hyp:capacity} and \ref{hyp:pri})}

Next, we examines the roles of multi-step returns (bootstrapping), network capacity (function approximation), and prioritisation (off-policy state distribution).
The plots in Figure \ref{fig:violin_triad} show the fraction of runs that exhibit soft-divergence, for each of the four bootstrap targets, as a function of bootstrap length, network size, and prioritisation. In the appendix, we additionally report the full distributions, across all experiments, of the maximal action values.

Figure~\ref{fig:violin_triad}a shows the effect of the bootstrap length $n$.  For all four bootstrap methods, there is a clear trend that a longer bootstrap length reduces the prevalence of instabilities. One-step Q-learning ($n=1$) exhibits soft divergences on 94\% of the runs. With $n=10$, this reduced to 21\%. The other bootstrap types also show clear correlations. This provides strong support for Hypothesis~\ref{hyp:steps}.

Figure~\ref{fig:violin_triad}b shows the effect of network capacity. The results ran counter to Hypothesis~\ref{hyp:capacity}. For Q-learning: small networks exhibited less instabilities ($53\%$) than larger network ($67\%$). The trend is less clear for more stable update rules; double Q-learning has a consistent soft-divergence rate of about $~10\%$ across network capacities. As results in Section \ref{fig:performance} show, despite a possibly mildly decreased stability, the control performance with the larger networks was higher.

Figure \ref{fig:violin_triad}c shows a consistent impact of changing the state distribution, through prioritised replay, on the soft-divergence rate. For Q-learning the rate increased from $52\%$ to $77\%$. For the more stable double Q-learning update the rate grows from $2\%$ to $23\%$. Furthermore, the use of importance sampling corrections has a measurable impact on the degree of instability in the value estimates. This is especially true for stronger prioritisation ($\alpha=1$ and $\alpha=2$). Without importance sampling correction (marked `UP' in Figure \ref{fig:violin_triad}c), the fraction of diverging runs is up to $10\%$ higher than with importance sampling correction (`P').
These results provide strong support for Hypothesis~\ref{hyp:pri}.

\subsection{Agent performance}\label{fig:performance}

\begin{figure*}
  \centering
  \includegraphics[height=0.32\linewidth]{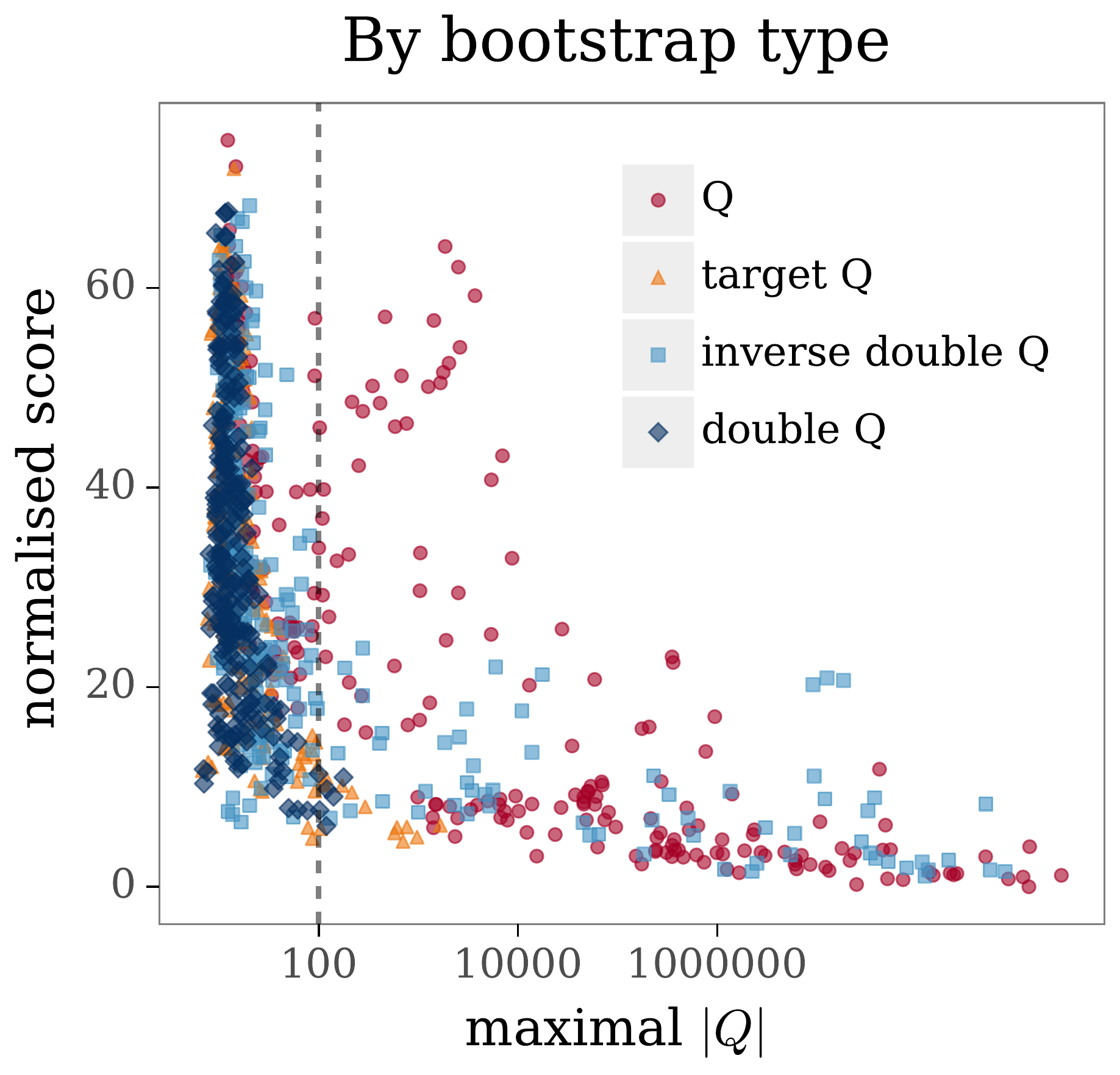}
  \includegraphics[height=0.32\linewidth]{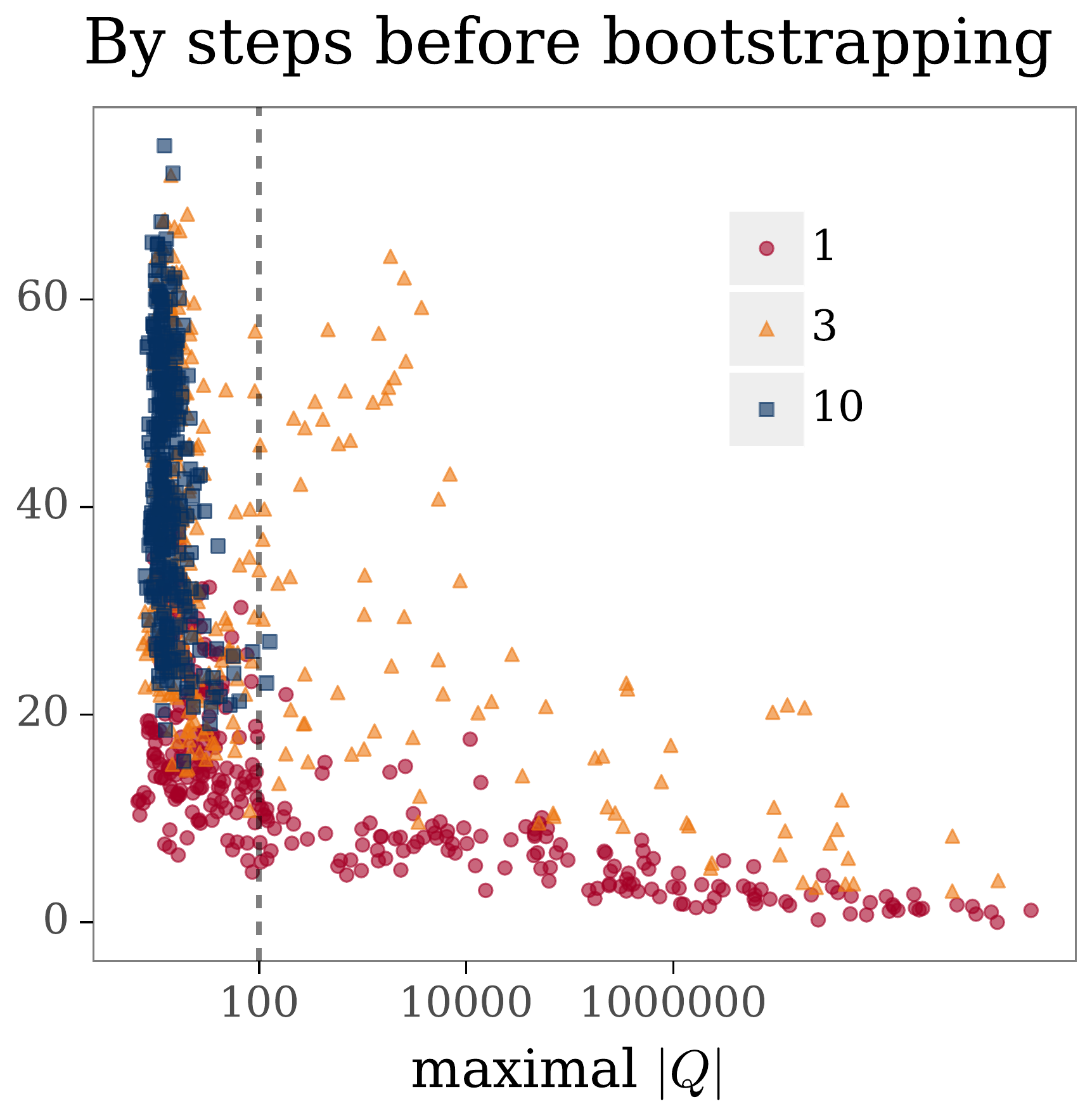}
  \includegraphics[height=0.32\linewidth]{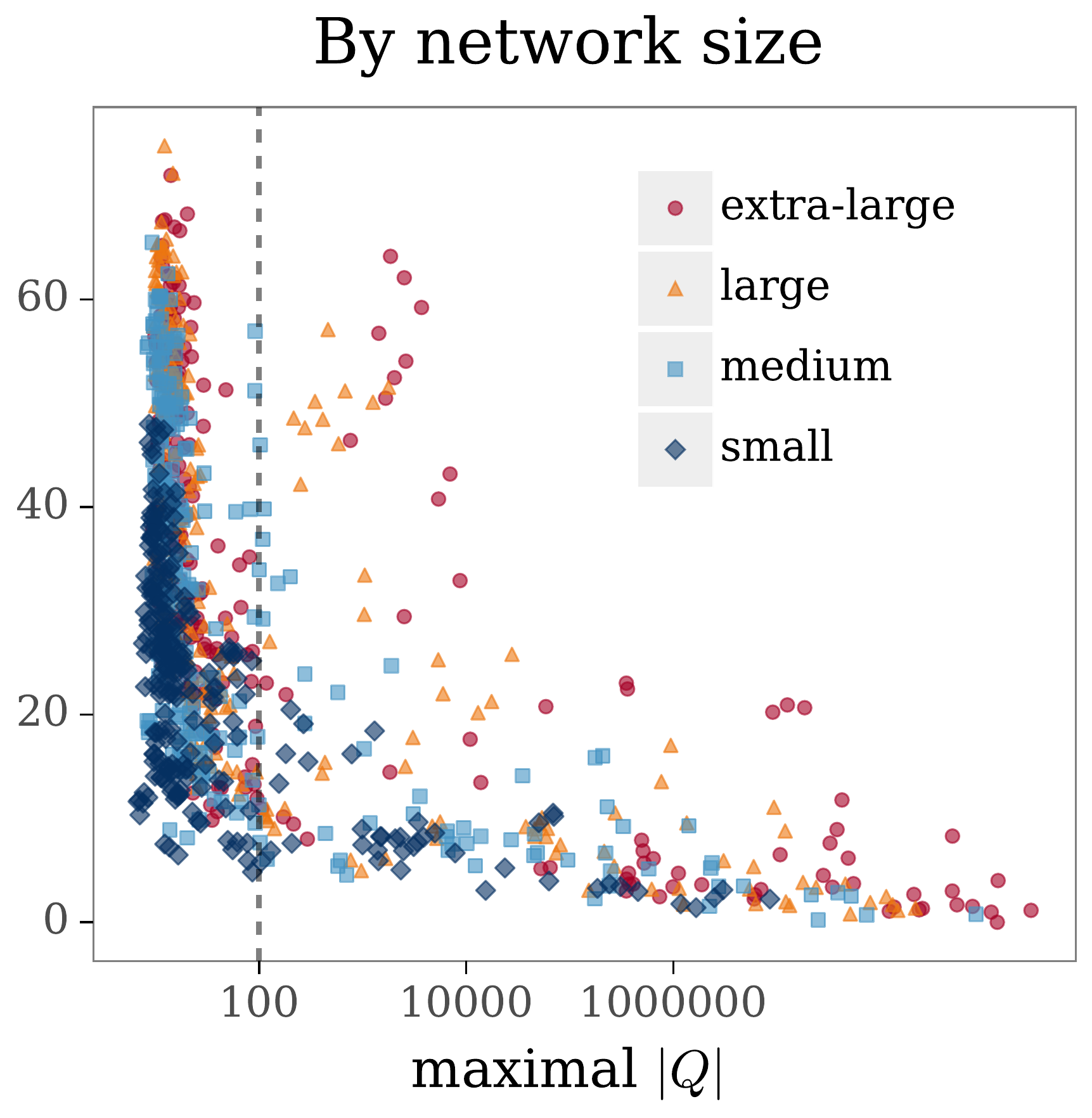}
  \includegraphics[height=0.32\linewidth]{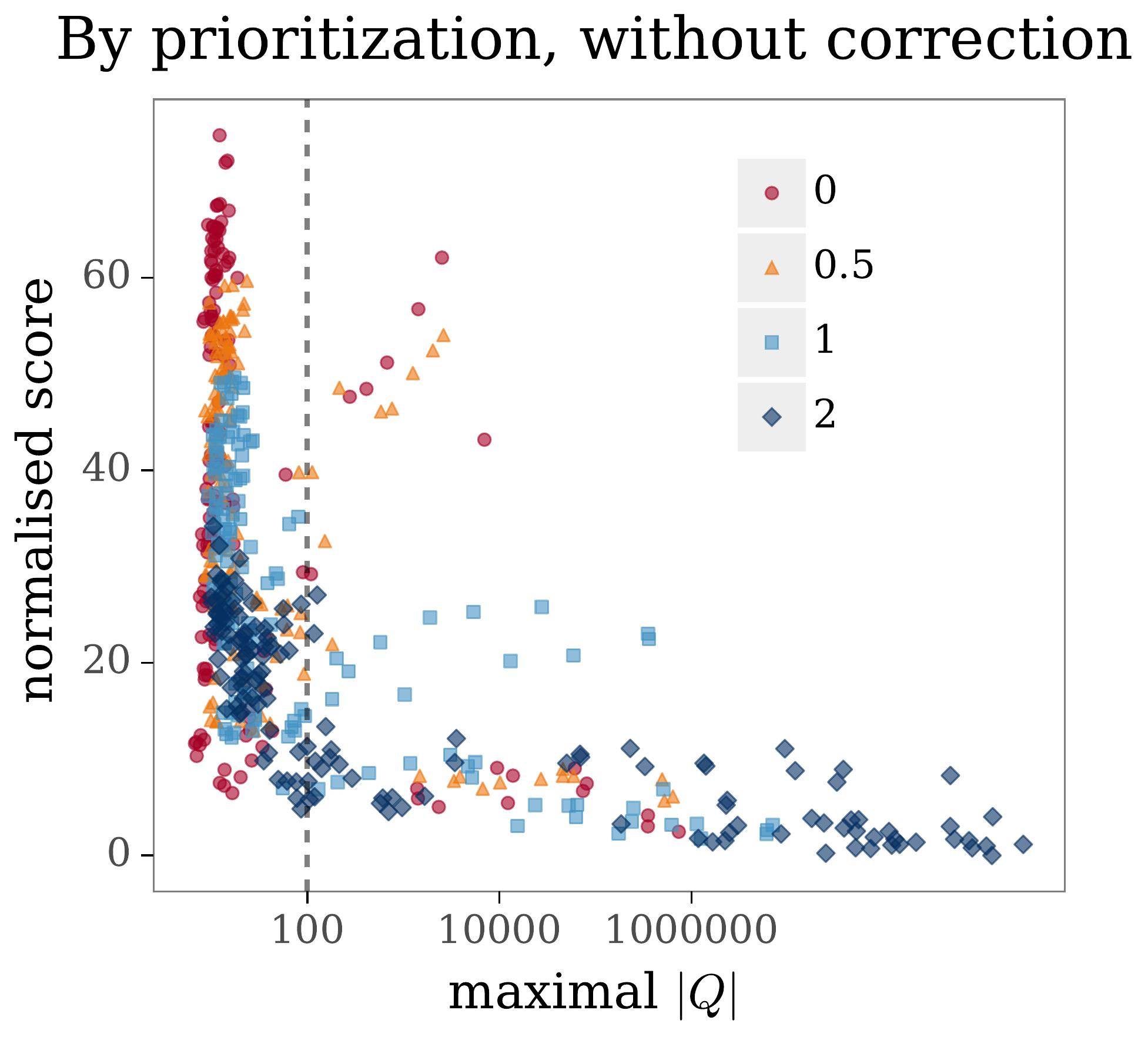}
  \includegraphics[height=0.32\linewidth]{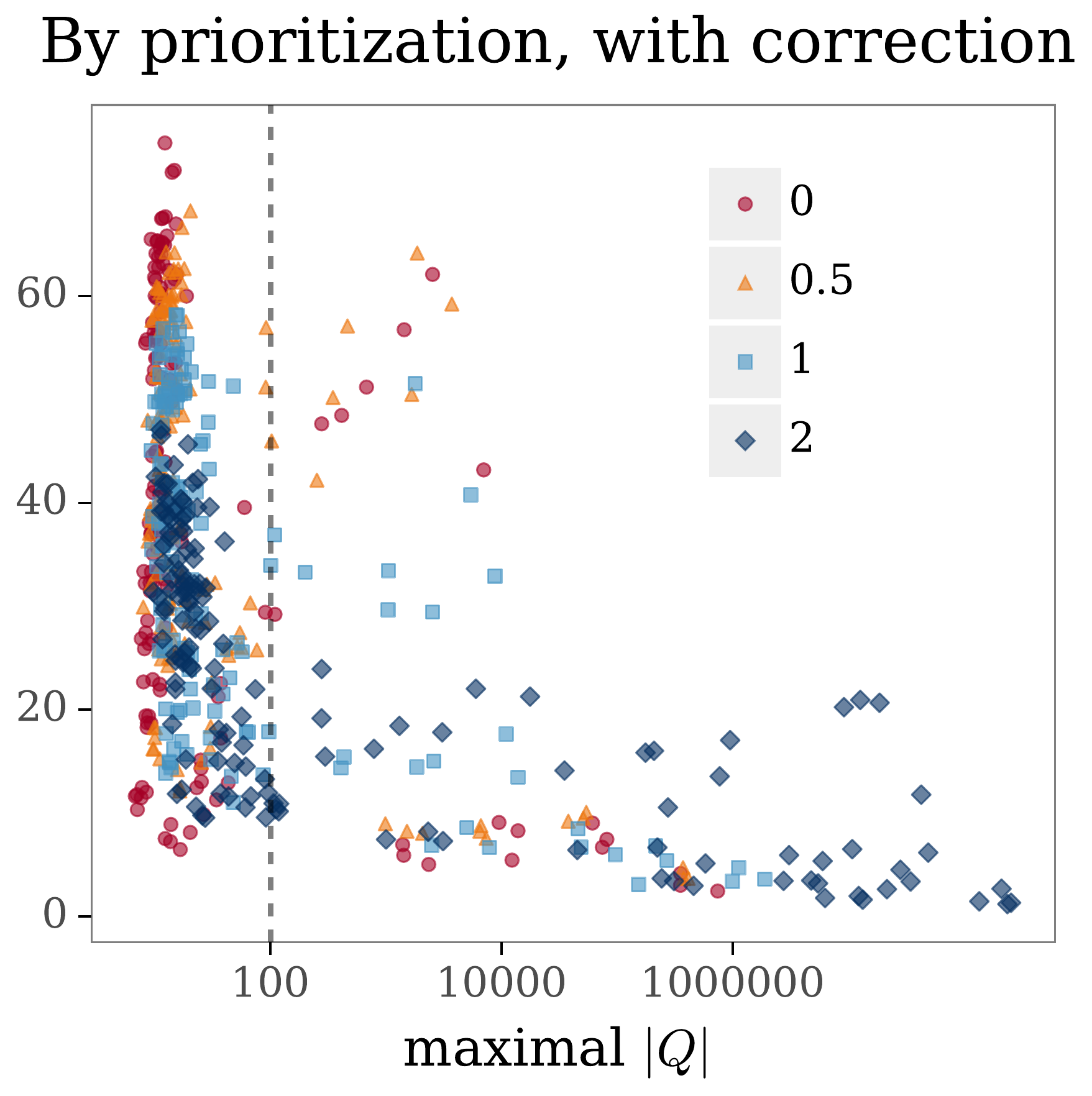}
  \caption{In these plots, each point's coordinates correspond to the median maximal value estimate and median human normalized performance, aggregated across the 57 games. The top row shows the full data-set, labelled and coloured according to different algorithmic choices. The second row shows the data split according to with (right) and without (left) importance sampling corrections, coloured according to the prioritisation parameter $\alpha$. The dashed vertical line denotes the maximum possible value of $100$. We see that in general, unrealistic value estimates correspond to poor performance.}
  \label{scatter}
\end{figure*}

We now investigate whether instabilities introduced by the deadly triad affect performance.
We look at how the performance of the agent varies across the $336$ different hyper-parameter configurations and $3$ replications. We measure control performance in terms of median human-normalised score~\citep{vanHasselt:2016} across 57 games. Figure \ref{scatter} shows the results for all parameters and replications, with performance on the $y$-axis and maximum value estimates on the $x$-axis. Thi

We see that soft-divergence (unrealistic value estimation) correlates with poor control performance. Experiments that did not exhibit soft-divergences (to the left of the dashed vertical line) more often have higher control performance than experiments that were affected by soft-divergence (to the right of the dashed vertical line), which exhibit both low performance and unrealistic value estimates.

In the top-left plot of Figure \ref{scatter}, the data is split by the bootstrap type. Q-learning and, to a lesser degree, inverse double Q-learning  soft-diverge most often, and then tend to perform much worse. Some Q-learning runs exhibit both unrealistic values (e.g., around 1,000) as well as reasonably good performance which indicates that in these cases Q-learning exhibits an overestimation bias, but that the ranking of the action values is roughly preserved.

In the top-center plot of Figure \ref{scatter}, we split the data according to the bootstrap length. The resulting pattern is clear: longer multi-step returns ($n=10$) correspond to fewer instabilities and better performance. With $n=3$ soft divergence is more common (especially for Q-learning). Finally, for $n=1$, we observe many unrealistic value estimates combined with poor performance.

A different pattern emerges when we look at network capacity (Figure \ref{scatter}, top-right). On one hand, the best performing experiments (top-left corner of the plot) use the bigger network architectures (labelled \textit{large}, and \textit{extra-large}). On the other hand, perhaps counter-intuitively, the largest networks also have the highest number of unrealistically high value estimates, at least for Q-learning.

Finally, the bottom row in Figure \ref{scatter} highlights the role of prioritisation. The two plots correspond to prioritisation with (right) and without (left) importance sampling corrections. The runs are coloured according to the prioritisation parameter $\alpha$. Both plots include uniform sampling ($\alpha=0$), in which importance sampling corrections have no effect. Prioritising too heavily ($\alpha=1$ or $\alpha=2$) correlates both with unreasonably high value estimates as well as reduced performance.

\section{Discussion}

Our empirical study of divergence in deep reinforcement learning revealed some novel and potentially surprising results.
Action value estimates can commonly exhibit exponential initial growth, and still subsequently recover to plausible magnitudes.   This property is mirrored in our extension to the Tsitsiklis and Van Roy example, and runs counter to a common perspective that deep neural networks and reinforcement learning are usually unstable.
Our hypotheses for modulating divergence with multi-step returns and different prioritisations were supported by the results, though the interaction of network capacity with the other components of the triad is more nuanced than we had initially hypothesized.
A key result is that the instabilities caused by the deadly triad interact with statistical estimation issues induced by the bootstrap method used.  As a result, the instabilities commonly observed in the standard $\epsilon$-greedy regime of deep Q-learning can be greatly reduced by bootstrapping on a separate network and by reducing the overestimation bias. 
These alternatives to the basic Q-learning updates do not, however, fully resolve the issues caused by the deadly triad. The continuing relevance of the deadly triad for divergence was shown with the increase of instabilities under strong prioritisation of the updates---corresponding to a strongly off-policy distribution. 
Turning to control, we found that this space of methods for modulating the divergences can also boost early performance. In our experiments, there were strong performance benefits from longer multi-step returns and from larger networks. Interestingly, while longer multi-step returns also yielded fewer unrealistically high values, larger networks resulted in \emph{more} instabilities, except when double Q-learning was used. We believe that the general learning dynamics and interactions between (soft)-divergence and control performance could benefit from further study.

\bibliography{bib}
\bibliographystyle{abbrvnat}

\newpage
\appendix

\section{Experiment details}
We use the same pre-processsing as DQN, including downsampling, greyscaling, frame-stacking of input images, 4 action repeatitions, reward clipping at $[-1, 1]$, and a discount factor of $0.99$.  

We used $\epsilon$-greedy action selection with $\epsilon=0.01$, a minibatch size of 32, and a replay buffer with the capacity for 1M transitions. We do not sample from the replay buffer until it is at least 20\% full and sample once every 4 agent steps (i.e. 16 environment frames) thereafter. We used a step size of $0.0001$ with the Adam optimizer~\citep{Kingma:2015}. When target networks were used for bootstrapping, the target network was updated every 2500 agent steps (i.e. 10,000 frames).  

The experiments were run with the full selection of 57 Atari games that has become standard in the deep RL literature  ~\citep{vanHasselt:2016, Wang:2016, Schaul:2016,
Horgan:2018}. Each configuration was run three times, with different random seeds.

In all our experiments we used fairly conventional network architecture. All networks use two convolutional layers followed by a fully connected hidden layer, and a linear output layer with as many outputs as there are actions in the game. All hidden layers are followed by a ReLU nonlinearity. The convolutional layers have respectively kernel size [$8 \times 8$, $4 \times 4$], and stride [$4, 2$]. The number of channels and hidden units in convolutional and fully connected layers are listed in the Table below.

\begin{table}[h]
\centering
\small
\begin{tabular}{l|rrr}

Network & channels (1) & channels (2) & hidden units \\
\hline
small       & 4  & 8  & 64  \\
medium      & 8  & 16 & 128 \\
large       & 16 & 32 & 256 \\
extra-large & 32 & 64 & 512 \\

\end{tabular}
\end{table}

\section{Training curve at early stages}

\begin{figure*}[h!]
    \centering
    \begin{subfigure}[t]{0.49\textwidth}    
        \centering
        \includegraphics[width=\linewidth]{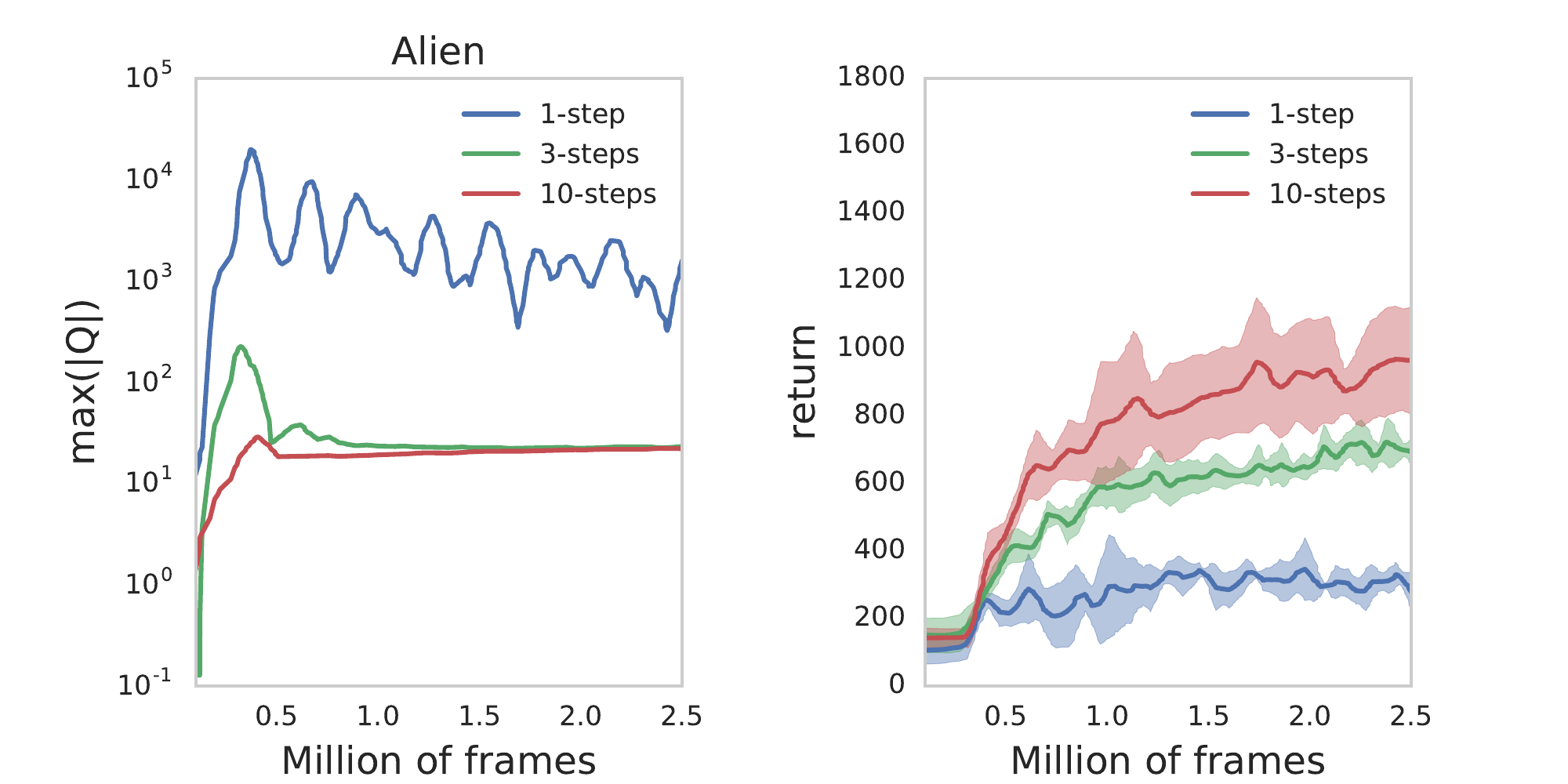}
        \caption{Early training on Alien \\comparing multi-step returns.}
    \end{subfigure}
    \begin{subfigure}[t]{0.49\textwidth}        
        \centering
 \includegraphics[width=\linewidth]{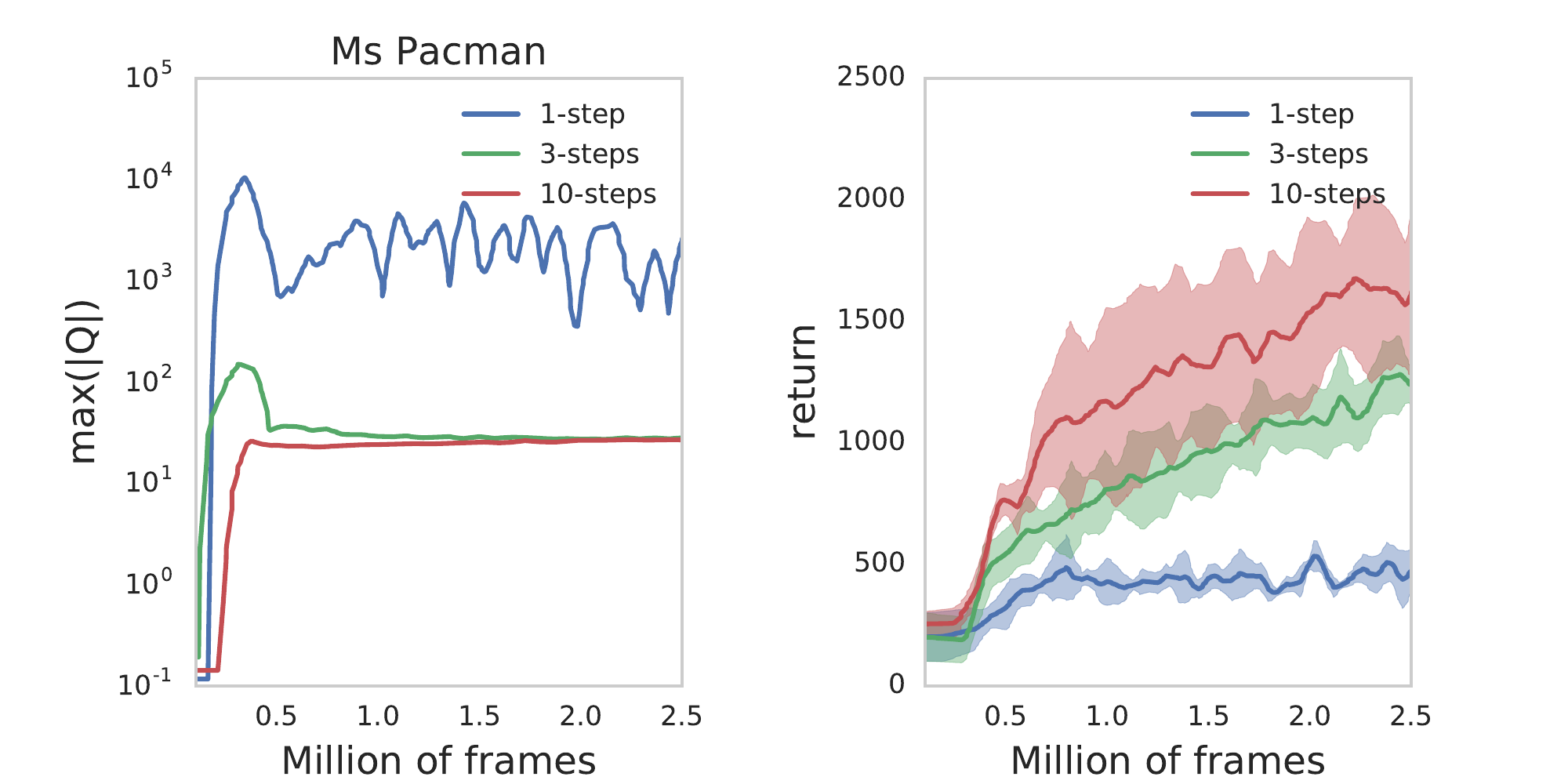}
        \caption{Early training on Ms Pacman \\comparing multi-step returns.}
    \end{subfigure}
    \\
    \begin{subfigure}[t]{0.49\textwidth}    
        \centering
        \includegraphics[width=\linewidth]{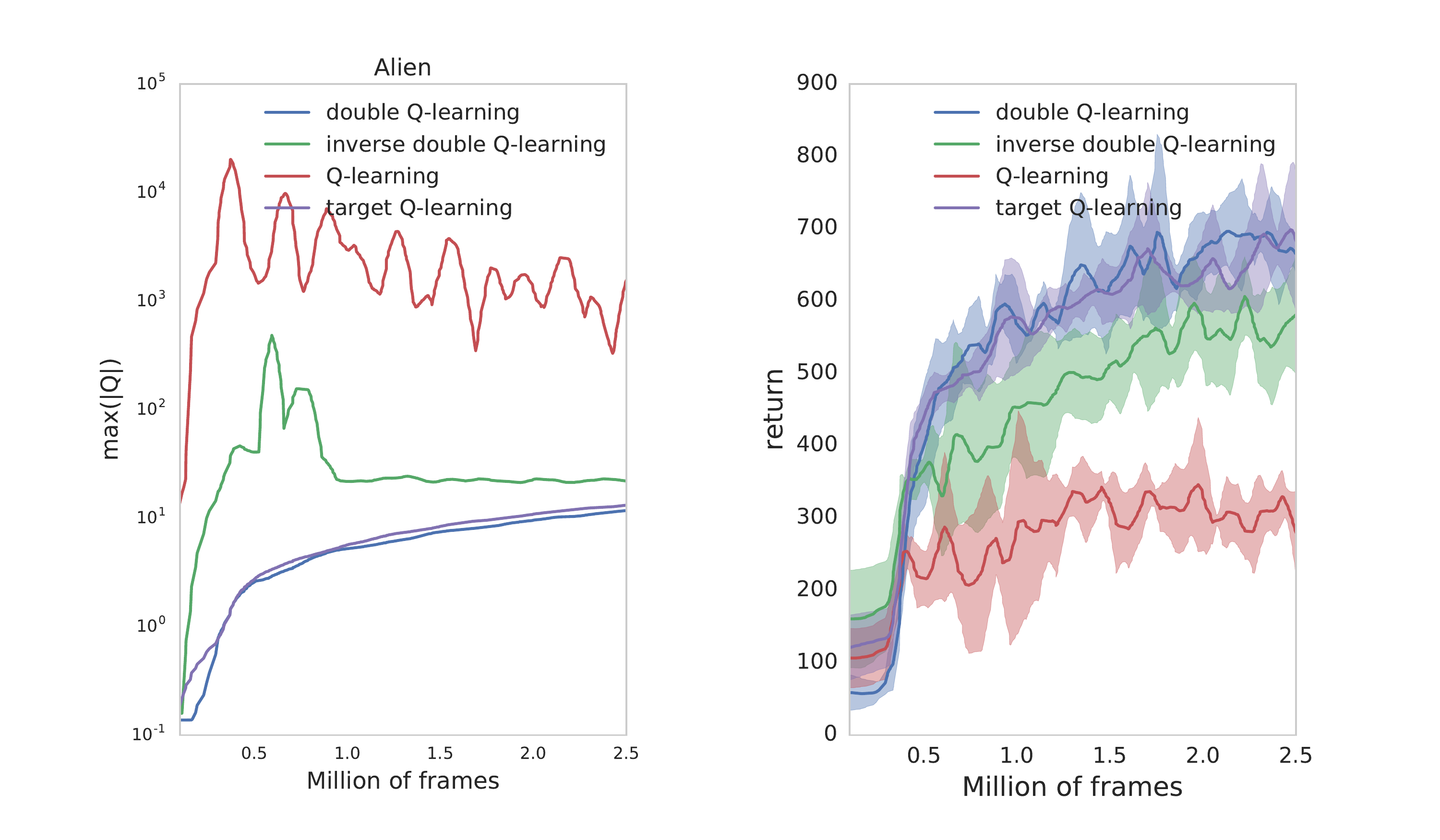}
        \caption{Early training on Alien \\comparing bootstrapping type.}
    \end{subfigure}
    \begin{subfigure}[t]{0.49\textwidth}       
            \centering
 \includegraphics[width=\linewidth]{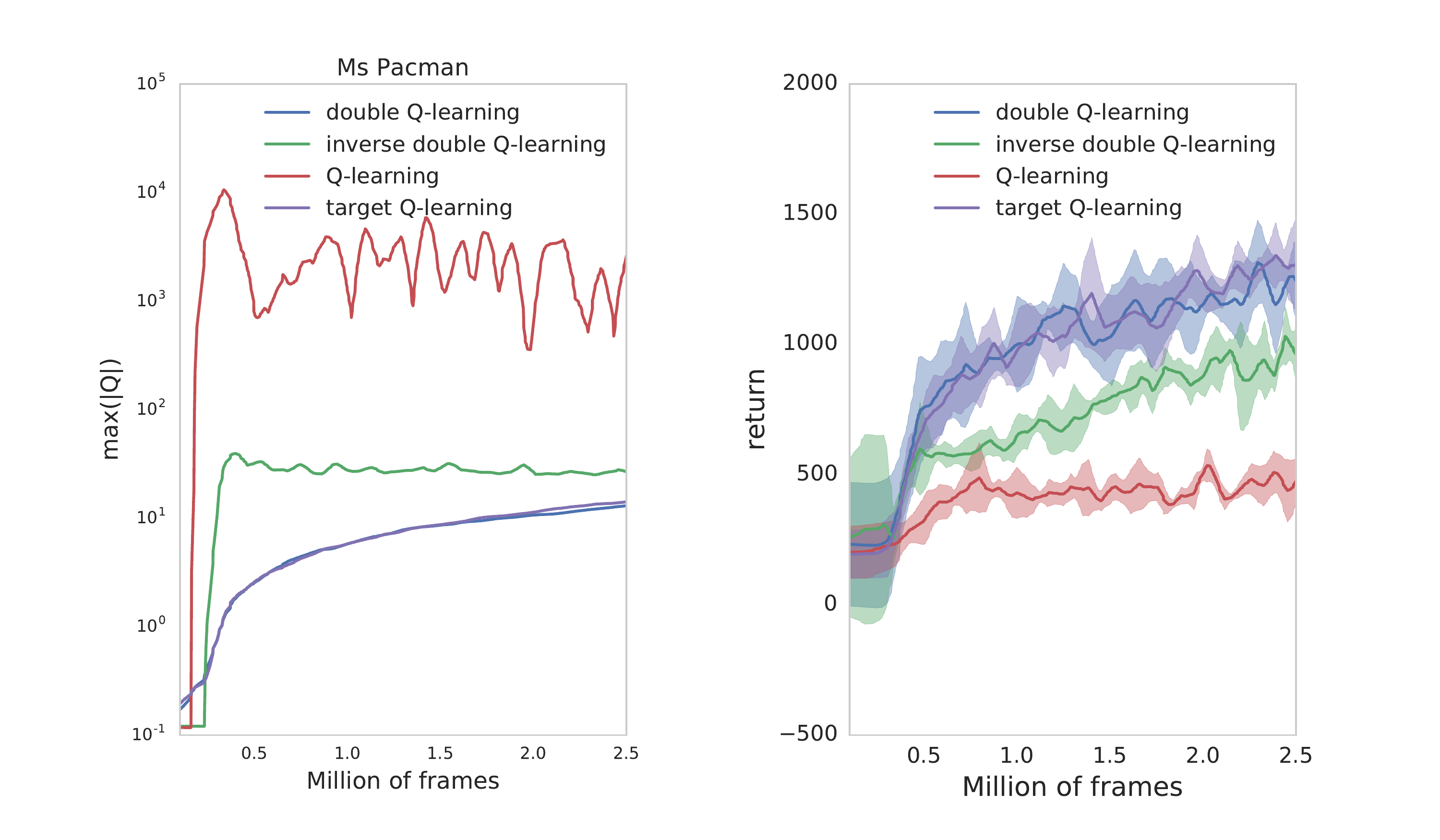}
        \caption{Early training on Ms Pacman \\comparing bootstrapping type.}
    \end{subfigure}
    \caption{\textbf{Comparing maximal absolute action-values and performance} Each figure plots the highest absolute action values and the mean episode return as a function of environment frames. Each slice is measured on an interval of roughly 50K--100K frames. The plots show different Atari games, Alien (left) and Ms Pacman (right), and different components - multi-step returns (top), bootstrapping type(bottom). We can observe the correlation between high action values and low returns.}\label{fig:plot_perf}
\end{figure*}

\newpage
\section{Visualising overestimation distribution}

\begin{figure*}[h!]
    \centering
    \begin{subfigure}[t]{\textwidth}    
        \centering
        \includegraphics[width=\linewidth]{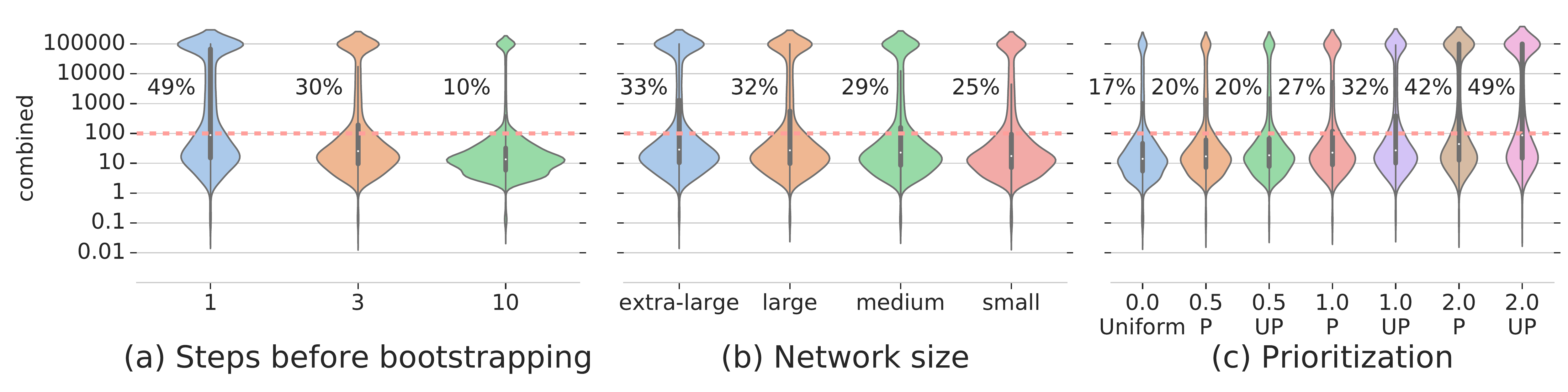}
        \caption{Across all bootstrap types.}
    \end{subfigure}
    \begin{subfigure}[t]{\textwidth}    
        \centering
        \includegraphics[width=\linewidth]{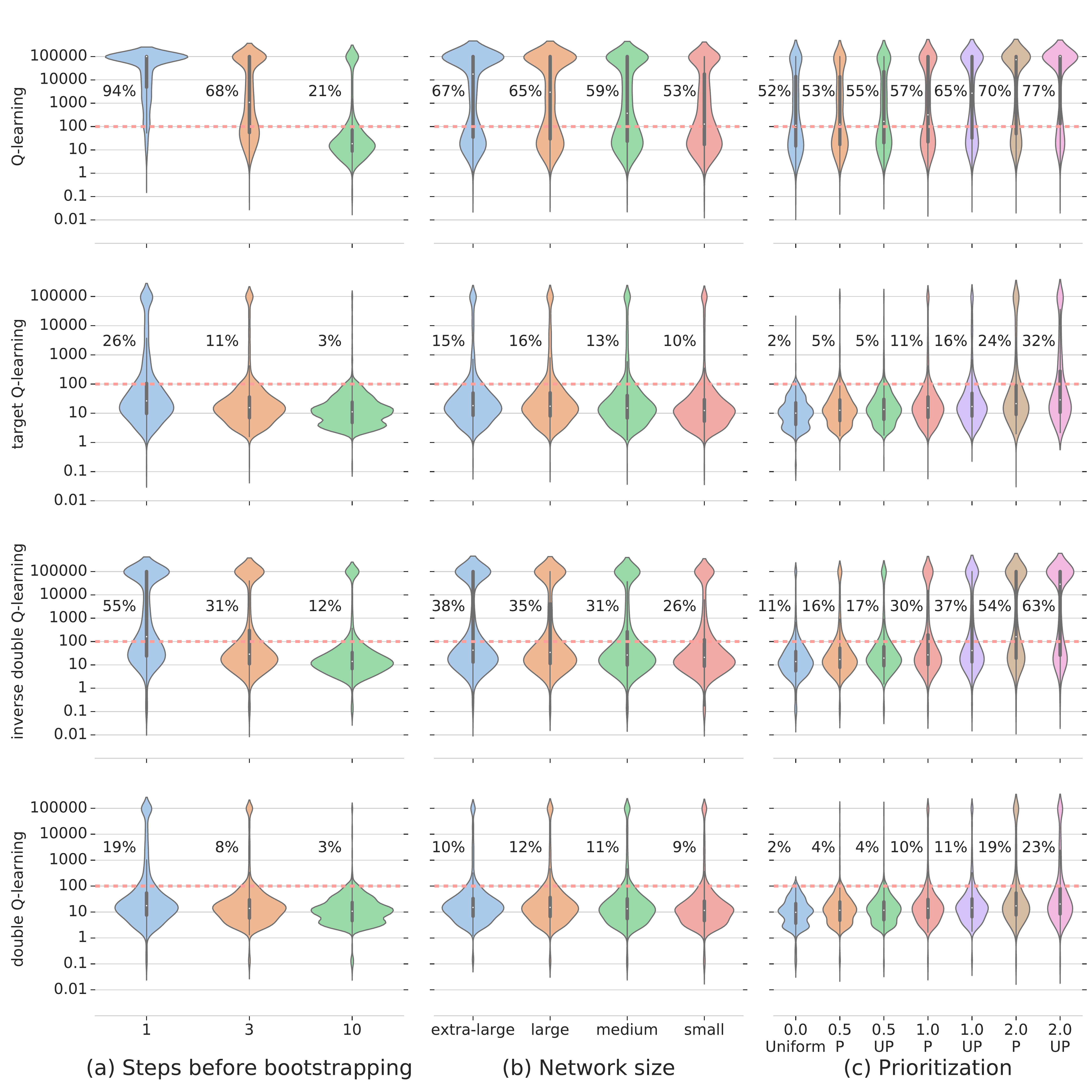}
        \caption{Split by bootstrap type.}
    \end{subfigure}
    \caption{\textbf{Maximal absolute action-values distributions} Maximal action-values for different splits of the experiments. The fraction of soft-diverging runs (mass above the dashed line) for each split is next to the corresponding violin plot. The top-most row contains the combined experiments, and the bottom four rows contain data from the different bootstrap types separately. We label the prioritisation column according to the $\alpha$ value, and
    whether the data is with or without importance sampling correction (`P` and `UP`, respectively).} 
    \label{fig:big_violin}. 
\end{figure*}

\newpage
\section{Scores breakdown on individual games}

\begin{table}[h]
\centering
\small
\linespread{0.8}\selectfont\centering
\begin{tabular}{l|r|r}
        \textbf{Atari game} & \textbf{normalised score} &  \textbf{episode return} \\
\hline
             alien &            22.2\% &               1590.6 \\
            amidar &            35.8\% &                554.6 \\
           assault &           250.1\% &               1238.9 \\
           asterix &            25.8\% &               2103.7 \\
         asteroids &             0.8\% &                999.5 \\
          atlantis &          5719.0\% &             797776.2 \\
        bank\_heist &           125.2\% &                803.6 \\
       battle\_zone &            66.5\% &              22769.6 \\
        beam\_rider &            17.8\% &               2966.7 \\
           berzerk &            20.7\% &                560.6 \\
           bowling &            35.5\% &                 66.8 \\
            boxing &           521.6\% &                 49.7 \\
          breakout &            91.5\% &                 25.7 \\
         centipede &             7.0\% &               2668.5 \\
   chopper\_command &            26.0\% &               2919.6 \\
     crazy\_climber &           354.7\% &              88417.5 \\
          defender &            57.8\% &               9479.9 \\
      demon\_attack &            13.9\% &                610.5 \\
       double\_dunk &           199.1\% &                -10.2 \\
            enduro &           144.9\% &               1072.4 \\
     fishing\_derby &            77.8\% &                -16.4 \\
           freeway &           126.5\% &                 32.4 \\
         frostbite &            75.0\% &               3169.2 \\
            gopher &           320.2\% &               6832.1 \\
          gravitar &            26.1\% &                941.5 \\
              hero &            42.7\% &              11618.0 \\
        ice\_hockey &            53.4\% &                 -5.0 \\
         jamesbond &           122.2\% &                444.0 \\
          kangaroo &           419.3\% &              11318.9 \\
             krull &          1194.2\% &               7701.5 \\
    kung\_fu\_master &           111.4\% &              23121.7 \\
 montezuma\_revenge &             0.0\% &                  1.0 \\
         ms\_pacman &            14.9\% &               2558.2 \\
    name\_this\_game &            95.1\% &               6575.1 \\
           phoenix &            62.7\% &               4477.3 \\
           pitfall &             3.6\% &                 -5.3 \\
              pong &           113.2\% &                 20.3 \\
       private\_eye &             0.8\% &                569.5 \\
             qbert &            87.3\% &              10567.5 \\
         riverraid &            43.0\% &               6945.8 \\
       road\_runner &           507.9\% &              34888.7 \\
          robotank &           515.0\% &                 36.7 \\
          seaquest &             5.1\% &               2144.8 \\
            skiing &            27.4\% &             -13417.5 \\
           solaris &             6.7\% &               1891.2 \\
    space\_invaders &            30.1\% &                543.7 \\
       star\_gunner &            13.2\% &               1832.5 \\
          surround &            11.4\% &                 -8.2 \\
            tennis &           138.7\% &                 -0.1 \\
        time\_pilot &            97.7\% &               5602.7 \\
         tutankham &           110.9\% &                152.1 \\
         up\_n\_down &           220.2\% &              21153.2 \\
           venture &           101.0\% &               1049.3 \\
     video\_pinball &          1110.4\% &              23094.7 \\
     wizard\_of\_wor &            74.7\% &               3544.9 \\
      yars\_revenge &            55.1\% &              27374.5 \\
            zaxxon &            85.1\% &               7189.7 \\
\end{tabular}
\vspace{13pt}
\caption{\textbf{Score breakdown on individual games}. For each game we report the raw and normalized score for the best parameter configuration after 20M frames. This is selected by taking for each game the highest mean human normalized return over 50 logging episodes, then computing the median of these scores across the 57 Atari games, and finally averaging over 3 runs of each experiment. The corresponding parameters where 10-step return, large network, Q-learning bootstrapping and uniform replay. For each game we report both raw and normalized returns}
\end{table}

\end{document}